# VTrackIt: A Synthetic Self-Driving Dataset with Infrastructure and Pooled Vehicle Information

Mayuresh Savargaonkar[1], *Student Member, IEEE*, Abdallah Chehade[1], *Member, IEEE*

*Abstract*—**Artificial intelligence solutions for Autonomous Vehicles (AVs) have been developed using publicly available datasets such as Argoverse, ApolloScape, Level5, and NuScenes. One major limitation of these datasets is the absence of infrastructure and/or pooled vehicle information like lane line type, vehicle speed, traffic signs, and intersections. Such information is necessary and not complementary to eliminating high-risk edge cases. The rapid advancements in Vehicle-to-Infrastructure and Vehicle-to-Vehicle technologies show promise that infrastructure and pooled vehicle information will soon be accessible in near real-time. Taking a leap in the future, we introduce the first comprehensive synthetic dataset with intelligent infrastructure and pooled vehicle information for advancing the next generation of AVs, named VTrackIt. We also introduce the first deep learning model (InfraGAN) for trajectory predictions that considers such information. Our experiments with InfraGAN show that the comprehensive information offered by VTrackIt reduces the number of high-risk edge cases. The VTrackIt dataset is publicly available upon request under the Creative Commons CC BY-NC-SA 4.0 license at https://vtrackit.irda.club.**

## I. INTRODUCTION

With the advancements in electronics and communication technologies, there is clear progress toward near real-time Vehicle-to-Infrastructure (V2I) and Vehicle-to-Vehicle (V2V) data transfer [1]. Such data transfer technologies provide an unprecedented opportunity for increased safety and reliability of Autonomous Vehicles (AVs) even in adverse situations [2], [3]. Examples of V2V data include speed, heading, and pedal positions. Examples of V2I data include red light status, speed limits, and lane data [4]. Vehicle-to-Everything (V2X) encompasses both V2V and V2I data [5]–[7]. Multiple research efforts have explored the potential of integrating intelligent infrastructure and connected vehicle data to develop individual AV applications [5]–[10]. A detailed review by Jeong *et al.* [11] summarizes many of these research efforts and explains the vision of smart cities. It is evident that V2X will be the next transformative step toward building safe, resilient, robust, and reliable AVs.

Unfortunately, broad availability of public datasets with infrastructure and pooled vehicle information is limited to large-scale industrial efforts in the form of proprietary collected data [12]. This limits the progress within the self-driving community to develop futuristic models that leverage concepts of smart cities for AV applications. To address this research gap, we introduce a synthetically generated self-driving dataset named VTrackIt. The name VTrackIt reflects tracking infrastructure and surrounding vehicle information in real-time. VTrackIt aims to attract the attention of the self-driving community to integrate infrastructure and pooled vehicle information for safer and more reliable AV applications. The VTrackIt [13] dataset is publicly available for use upon request.

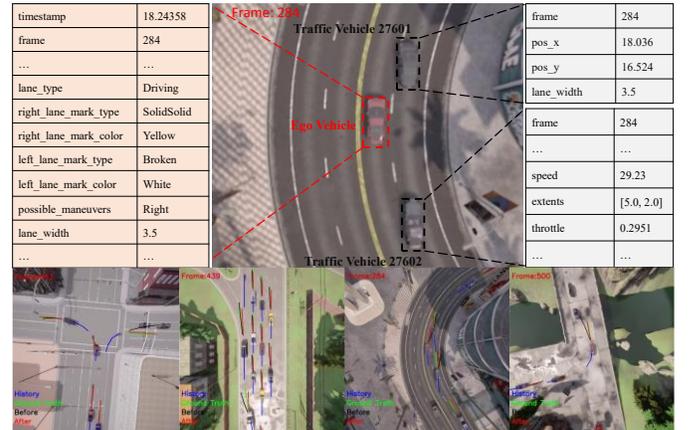

Fig. 1. An overview of VTrackIt with infrastructure and pooled vehicle information. Observed trajectories are labeled as 'History' and future trajectories are labeled as 'Ground Truth'. Predicted trajectories using state-of-the-art benchmark (without infrastructure information) are labeled as 'Before.' Predicted trajectories using VTrackIt are labeled as 'After.'

VTrackIt is inspired by existing datasets like Level5 [12], NuScenes [14], Argoverse [15], ApolloScape [16], Waymo [17], AIODrive [18], and others. None provide a comprehensive vision of integrating concepts of intelligent infrastructure and smart cities for the safer deployment of AVs. Specifically, Argoverse, NuScenes, AIODrive, and ApolloScape do not log relevant infrastructure information like lane annotation, lane widths, speed limits, stop signs, and many more. One dataset that seems to record some infrastructure data is Level5; however, it fails to provide data from a pool of surrounding vehicles like speed, lateral and longitudinal accelerations, and pedal positions, among others. It is worth noting that some datasets offer semantic maps, but they are biased to sequences within their given Operational Design Domain (ODD) [12], [15], [17]. This bias makes it challenging to develop perception and other Deep Learning (DL) models that are generalizable beyond the training data. Taking a leap in the future, VTrackIt provides a more comprehensive vision of integrating infrastructure and surrounding vehicle information. We also validate the advantages of VTrackIt by developing a Generative Adversarial Network (GAN), called as InfraGAN for trajectory predictions. Our experiments with InfraGAN show that high-risk edge cases can be reduced by using VTrackIt. The contributions of this paper are multifold:

1. *A large-scale, synthetic, self-driving dataset with infrastructure and surrounding vehicle information.*

2. *A video database with high-resolution Birds-Eye-View (BEV) and 360° views of the ego vehicle to promote the development of semantic-map-free solutions.*

3. *A compelling case study for using VTrackIt. Specifically, the first GAN (InfraGAN) that integrates infrastructure and pooled vehicle information for trajectory predictions.*

[1]Mayuresh Savargaonkar, and Abdallah Chehade are with the Department of Industrial Manufacturing and Systems Engineering at the University of Michigan-Dearborn, MI 48128, USA.
(e-mail: mayuresh@umich.edu; achehade@umich.edu).

We acknowledge that synthetically generated datasets will not replace real-world datasets. However, VTrackIt and other synthetic datasets should still be used to explore opportunities and significant infrastructure and surrounding vehicle information variables that are presently hard to integrate or inaccessible in real-world datasets.

## II. Literature Review: Related Work

The following section reviews commonly used self-driving datasets developed and released for public use.

### A. Real-world Driving Datasets

NGSIM [19] is one of the earliest naturalistic self-driving datasets used to develop several Machine Learning (ML) solutions. Although the goal of this dataset was to study traffic flow theories, limited infrastructure information was provided like lane ids, lane section id, and lane direction. Also, some surrounding vehicle information is provided, including vehicle extents, class, speed, and accelerations were captured at 10 Frames-Per-Second (FPS). However, NGSM is limited by rigid location constraints using a small section on the southbound US-101 and Lankershim Boulevard in Los Angeles, CA, eastbound I-80 in Emeryville, CA, and Peachtree Street in Atlanta, Georgia. Huang *et al.* [16] proposed the ApolloScape dataset that logged over 2 hours of driving, including 103 scenarios with complex vehicle and pedestrian traffic flows captured at 2 FPS. However, ApolloScape does not include annotations for relevant infrastructure information. Chang *et al.* [15] proposed Argoverse, the first large-scale self-driving dataset with detailed semantic maps for the cities of Pittsburgh and Miami captured at 10 FPS. Although the Argoverse dataset logs about 320 hours of driving across 324k complex scenarios using a semantic HD map with lane center positions, it fails to explicitly annotate crucial infrastructure information such as lane annotations. Instead, it is assumed that the semantic maps encoded with lane centers and connectivity provide enough information. The Argoverse dataset is supplemented by rasterized maps to identify drivable areas and corresponding ground height information. This dataset represents traffic vehicles using centroids, and no other information is provided apart from their locations. The Argoverse dataset fails to offer detailed surrounding vehicle information. Houston *et al.* [15] proposed the Level5 dataset that logged 1,118 hours of driving across 170k scenes with some surrounding vehicle information captured at 10 FPS. Level5 offers semantic maps with no infrastructure information. A major limitation of Level5 is its bias to a single heavily trafficked route under rigid location constraints. Holger *et al.* [14] released the NuScenes dataset that logged over 15 hours of driving across 1000 scenarios captured at 2 FPS. The NuScenes dataset includes annotations for vehicle category, locations, extents, and yaw, for a pool of surrounding vehicles. However, only limited infrastructure information is provided. Another old dataset for developing DL solutions for AVs is the KITTI dataset [20] which logged 22 minutes of driving captured at 10 FPS. Osinski *et al.* [21] proposed the OpenDD dataset with 501 scenarios recorded at 30 FPS; however, again, with limited infrastructure information. Finally, IntentNet [22] is a dataset used to develop DL solutions for AV applications but is not publicly available.

### B. Synthetic Driving Datasets

To advance the development of DL algorithms for use in AVs, some work has also been done to collect synthetic data using state-of-art simulators such as CARLA [23]. For example, Weng *et al.* [6] released the AIODrive dataset. Although this dataset includes data from different synchronized sensors, it lacks infrastructure information and diverse road conditions and does not offer a BEV. Xu *et al.* [24] introduced the OPV2V dataset with 70 interesting scenarios recorded using various; however, it does not investigate use of infrastructure information. KITTI-CARLA [25] is another publicly available synthetic dataset that records information identical to the KITTI dataset over seven simulated scenarios at 10 FPS. Some other synthetic datasets [28], [29] have also been made publicly available using simulation-based environments in recent years. However, most of these datasets are purpose-driven and only record LiDAR and/or perception-based information. Table 1 presents an overview of all significant datasets used for the development of several modern AV applications. While multiple efforts collect limited information from surrounding vehicles, there is a clear gap of leveraging intelligent infrastructure information for most of the discussed driving datasets. Next, we introduce the VTrackIt dataset [13].

## III. The VTRACKIT Dataset

The VTrackIt dataset consists of 600 scenarios (360 training, 120 validation, and 120 test), each recorded for a maximum of 30 seconds (the duration of some scenarios where the ego vehicle had a crash may be less than 30 seconds.) Uniquely, VTrackIt provides explicit annotations for several V2I and V2V information variables for the ego vehicle and all other traffic vehicles within a 50-meter radius surrounding the ego vehicle. All scenarios are generated using the CARLA simulator 0.9.13 [23]. Although several state-of-art simulators such as Nvidia DriveSim[1], Autonovi-sim [26], and GTA-V [27] exist, these simulators are not open source and may limit the future extension of this dataset. Although simulators such as TORCS [28] and AirSim [29] are publicly available, they offer limited sensor suites and lack the photorealistic rendering capabilities needed to train modern perception-based DL algorithms. Due to its seamless integration with Unreal Engine[2], CARLA offers several advantages over other mentioned simulators, such as (i) enhanced realism, (ii) sophisticated vehicle dynamics, (iii) map customizations, and (iv) easy and realistic, traffic and ego vehicle customizations.

TABLE 1
COMPARISON OF MAJOR DRIVING DATASETS USED FOR AV APPLICATIONS

| Dataset | Size | Scenarios | Maps | Trajectories | Synthetic | Frequency (FPS) | BEV | V2V Information | V2I Information | Diverse Road Conditions | Low Light Conditions |
|---|---|---|---|---|---|---|---|---|---|---|---|
| NGSIM [19] | 1.5h | - | Sections of US-101, and I-80 highways | ✓ | - | 10 | - | ✓ | ✓ (Partial) | - | - |
| KITTI [20] | 6h | 50 | Karlsruhe | ✓ | - | 10 | ✓ | ✓ | - | - | - |
| ApolloScape [16] | 2h | 103; 60s each | China | ✓ | - | 2 | - | ✓ | - | - | ✓ |
| Argoverse [15] | 320h | 324k; 6s each | Pittsburgh, Pa /Miami, FL | ✓ | - | 10 | ✓ (Semantic) | - | ✓ (Partial) | ✓ | ✓ |
| Level5 [12] | 1118h | 170k; 25s each | Single route in Palo Alto, CA | ✓ | - | 10 | ✓ (Aerial + Semantic) | - | ✓ (Partial) | - | - |
| NuScenes [14] | 5.5h | 1000; 20s each | Boston, Singapore | ✓ | - | 2 | ✓ (Semantic) | ✓ | - | ✓ | ✓ |
| AIODrive [18] | 2.8h | 100; 100 secs each | 8 Virtual maps with varying typologies | ✓ | ✓ | 10 | - | ✓ | - | ✓ | ✓ |
| OpenDD [21] | 62.7h | 501; 5-15 min each | 7 Round-abouts | ✓ | - | 30 | ✓ (Aerial + Semantic) | ✓ | - | - | - |
| **VTrackIt (Proposed)** | 5h | 600; 30s each | 6 virtual maps with varying typologies | ✓ | ✓ | 20 | ✓ (Aerial) | ✓ | ✓ | ✓ | ✓ |

The VTrackIt dataset is generated using the 'DirectSim' platform that will be released soon. The details of the VTrackIt dataset are provided in the following section.

*A. Sensing Package, Lane Annotations, and Aerial BEV*

To generate synthetic data for the VTrackIt dataset, we equip the ego vehicle (Tesla Model 3) in the simulation environment with four highly synchronized RGB cameras, one IMU, and one GNSS sensor. Specifications of all attached sensors are given in Table 2. Figure 2 visualizes the sensor positions and their orientations from the vehicle center as defined in CARLA. While the four RGB cameras with 120° Field-of-View (FoV) provide a 360° view around the ego vehicle, VTrackIt also provides an aerial BEV of the ego vehicle. All scenarios are captured and tagged at 20 FPS along with corresponding IMU and GNSS measurements.

TABLE 2
SENSING PACKAGE

| Sensor | Sensor Position Code | Location (in m) | Sensor Details |
|---|---|---|---|
| RGB Camera (Front) x 1 | a | X=0.9; Y=0; Z=1.44 | Forward facing stereo camera with 120° field-of-view recorded at 20 FPS. |
| RGB Camera (Right) x 1 | b | X=0.9; Y=1.2; Z=1.44 | |
| RGB Camera (Left) x 1 | c | X=0.9; Y= -1.2; Z=1.44 | |
| RGB Camera (Rear) x 1 | d | X=-0.9; Y=0; Z=1.44 | |
| IMU x 1 | e | X=0; Y=0; Z=0 | Inertial measurement sensor to record ego vehicle information at 20 FPS. |
| GNSS x 1 | e | X=0; Y=0; Z=0 | Global navigation sensor used to record ego vehicle pose information at 20 FPS. |

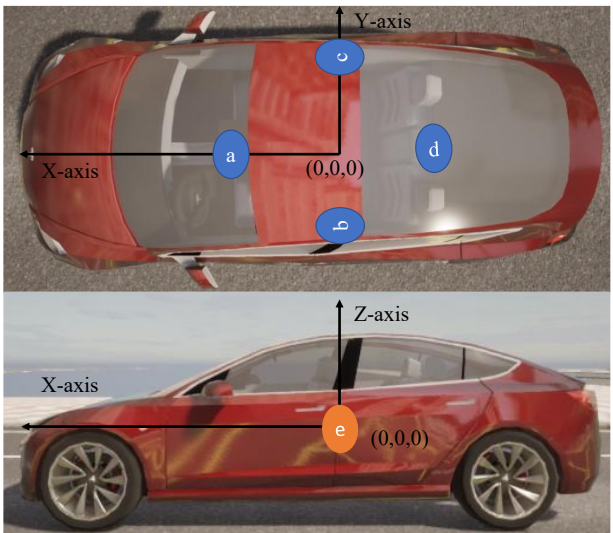

Fig. 2. Coordinate system for data recorded and locations of sensors with their respective sensor codes in the VTrackIt dataset.

In addition to the provided perception information, VTrackIt also includes annotations for vehicle information, including their locations, attributes, color, extents, heading, steer, throttle, and brake pedal positions, along with the relative position from the center of the ego vehicle to the center of the tracked traffic vehicle. Further, VTrackIt also annotates several infrastructure information variables, including lane line color and lane line type on both sides, possible maneuvers (lane restrictions), lane width, vehicle deviation from the centerline, and red-light status. Such data is provided for the ego vehicle and a pool of surrounding vehicles. Data from the surrounding vehicles can thus be used to train DL solutions for AVs in addition to the ego vehicle data. Table 3 provides detailed information on all annotated variables, data types, ranges, and measurement units. A brief description of each variable can also be seen in Table 3. Figure 3 shows a sample frame in a random scenario captured by all four RGB cameras giving the ego vehicle a 360° view of its surroundings. Note that each variable is annotated with a corresponding frame number that helps synchronize perception-based sensor information with other variables in the VTrackIt dataset.

TABLE 3
VARIABLES ANNOTATED AT EVERY RECORDED FRAME IN THE VTRACKIT DATASET

| Variable | Data Type | Description | Range / Possible Values | Unit of Measurement |
|---|---|---|---|---|
| timestamp | Float | Time stamp of the measurement | [0, inf) Increments by 0.05 | Seconds |
| frame | Integer | Frame number of measurement | [0, inf) Increments by 1 | int |
| actor_id | Integer | Unique id given to every actor in the scenario | [0, inf) | - |
| actor_type | String | Identifier given to distinguish ego vehicle from traffic vehicles | Ego / Traffic | - |
| attr | String | Classification of vehicle in CARLA standards | - | String |
| color | Tuple | RGB values of given actor | (0-255,0-255,0-255) | - |
| pos_x | Float | Global location of given actor along X-axis in cartesian co-ordinate system | [-inf, inf] | Meters |
| pos_y | Float | Global location of actor along Y-axis in cartesian co-ordinate system | [-inf, inf] | Meters |
| pos_z | Float | Global location of actor along Z-axis in cartesian co-ordinate system | [-inf, inf] | Meters |
| heading | Float | Global heading of actor relative to the map's true North | (0,360] | Degrees |
| extents | List | Actor length and width | [0, inf] | Meters |
| **speed** | Float | Actor speed | (0, inf] | KMPH |
| acceleration | List | Actor acceleration in X, and Y axes | [-inf, inf] | $m/s^2$ |
| **throttle** | Float | Throttle pedal position for given actor | (0,1) | - |
| **steer** | Float | Steer angle for given actor | (-1,1) | - |
| **brake** | Float | Brake pedal position given actor | (0,1) | - |
| **red_light** | Binary | Unique identifier that is set to '1' if vehicle is directly affected by a red light. | 0/1 | - |
| rel_angle | Float | Relative angle of a traffic vehicle measured from center of ego vehicle; for example vehicle front is 90° | (0,360] | degrees |
| rel_x | Float | Relative position of a traffic vehicle measured from center of ego vehicle along its y-axis; | (0,50) | Meters |
| rel_y | Float | Relative position of a traffic vehicle measured from center of ego vehicle along its x-axis | (0,50) | Meters |
| **lane_type** | String | Lane type affecting actor location | Driving / Junction / Shoulder | - |
| **right_lane_mark_type** | String | Right lane marking type affecting actor location | Solid / Broken / SolidSolid / NONE | - |
| **right_lane_mark_color** | String | Right lane marking color affecting given actor location | White / Yellow | - |
| **left_lane_mark_type** | String | Left lane marking type affecting actor location | Solid / Broken / SolidSolid / NONE | - |
| **left_lane_mark_color** | String | Left lane marking color affecting given actor location | White / Yellow | - |
| **possible_manuvers** | String | Permissible lane changes for given actor based on its location | Left / Right / Both / None | - |
| **lane_width** | Float | Width of driving lane based on given actor location | (0, inf] | Meters |
| **off_center** | Float | Deviations along lane center lines recorded in along given actor's X-axis. | (0, inf] | Meters |

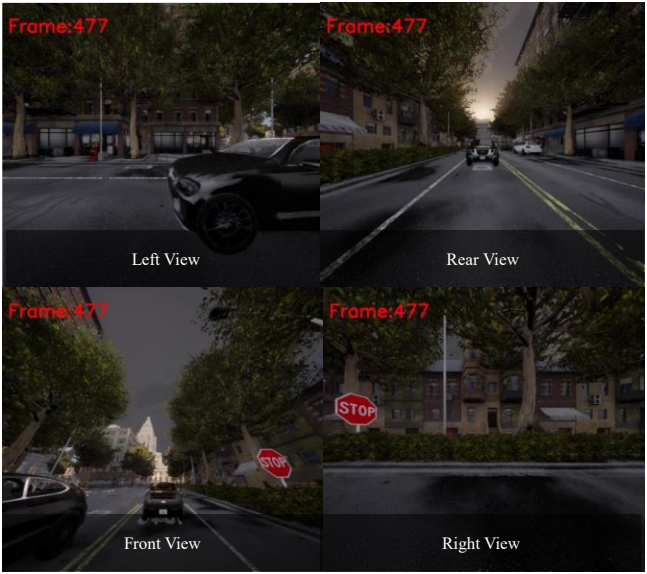

Fig. 3. 360° coverage of area surrounding the ego vehicle using four RGB cameras.

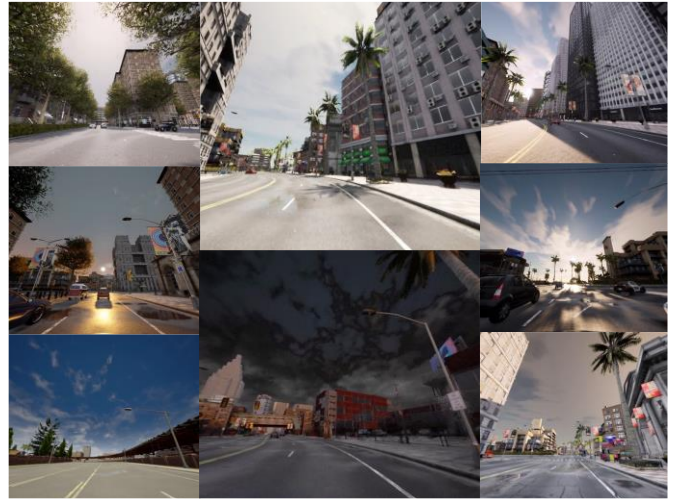

Fig. 4. Diverse set of weather and Typologys included in the VTrackIt dataset.

### B. Diversity in Weather and Typologies

A significant limitation of existing synthetic datasets is their lack of ability to replicate real-world driving conditions. Thus, it is of high interest to (i) simulate scenarios that depict close to real-life road typologies and (ii) simulate scenarios that adequately consider the interaction of vehicles with their surrounding environment. Furthermore, a simulated dataset such as VTrackIt must also include scenarios under varying operating conditions, including those induced by adverse weather. We thus randomize the weather in different maps in VTrackIt. Typically, we define and randomly sample weather from various weather conditions ranging from noon to sunset, clear to foggy, and from dry to wet. We also modify the road friction to closely mimic real-life conditions based on values given by Hall *et al.* [30] for wet weather scenarios. Figure 4 visualizes some of these weather conditions. Table 4 details various CARLA maps used in VTrackIt and their respective salient features.

### C. Traffic and Ego Vehicle Customizations

As mentioned earlier, interactions with the surrounding environment and traffic regulations play a significant role in developing DL solutions for AVs. Traffic is undoubtedly one aspect of it. However, no two vehicles are the same in real life because of the complex interaction between factors such as driver behavior and intrinsic differences in the driven vehicle, such as tire wear, vehicle age, and many more. Unfortunately, most synthetically generated datasets fail to intelligently model such vehicle attributes and assume similar attributes for all traffic actors. In their work, Weng *et al.* [18] briefly discuss and try to address this issue. In our case, we adopt a fuzzing strategy and randomize every traffic actor in every scenario by varying its (i) vehicle type, (ii) vehicle color, (iii) minimum following distance, (iv) maximum speed over/under speed limits, (v)

#### TABLE 4
#### MAP AND ROAD FEATURES

| Map Name | Location | Salient Road Features |
|---|---|---|
| Town01 | Low density urban (20-45 kmph ) | 3-way intersections, single lane roads, traffic lights, and stop signs. |
| Town03 | Urban-Highway (20-90kmph) | 5-lane intersections, roundabouts, uneven grades, tunnel, highways, traffic lights, and stop signs. |
| Town04 | Highway (20-90kmph) | Lane merges, highway ramps, and 4-way junctions. |
| Town05 | Hybrid (20-90 kmph) | Multiple highway lanes, urban lanes, and 4-way junctions. |
| Town06 | Highways (20-90 kmph) | Long highways, ramps, Michigan-left, round abouts, and lane merges. |
| Town10HD | Urban (20-45 kmph) | High fidelity road textures, parking lanes, stop signs, traffic lights, pedestrian crossings, and junctions. |

probability of ignoring other vehicles, and (vi) probability of ignoring traffic regulations, as defined in the CARLA documentation [23]. All of these variable values are sampled randomly from a uniform distribution for every variable, for every actor, in every scenario, based on real-life values extracted from naturalistic driving studies such as [31]–[34],[35]. Further, a few selected actors in every scenario are uniquely modified using out-of-distribution values to represent overly 'aggressive' and 'cautious' drivers. Such drivers are known to pose additional risks to the ego vehicle. Finally, to augment the realism of wet weather scenarios with low friction conditions, we modify the speed limits in that sequence based on naturalistic driving studies such as [36]–[38].

In our experiments, we use the same ego vehicle model in all scenarios to remain consistent with other real-life datasets. The simulator parameters for every scenario are sampled from predefined statistical distributions. Sampled parameters include safe distance, maximum speed over/under speed limits, probability of ignoring other vehicles, and probability of ignoring traffic regulations. The statistical distributions are designed based on the abovementioned datasets and NHTSA reports.

*D. Comparisons to Real-world Datasets*

Most simulator parameters are carefully designed to mimic real-world conditions in our work. We are specifically interested in replicating real-world traffic speed distributions and expanding beyond its right tail to encourage events that may lead to worst-case conditions such as crashes or near-crashes. Such scenarios are purposely excluded in real-world driving for obvious safety reasons; however, this introduces a bias toward safe scenarios. As shown in Fig. 5 (a) and (b), while the ApolloScape and Argoverse datasets contain many low-speed vehicles, the VTrackIt dataset expands the right-tail of the distribution by spawning more medium-high vehicles. The VTrackIt dataset thus capitalizes on the power of synthetic data generation tools to generate a wider distribution of traffic and ego vehicles that are likely to be encountered in a real-world scenario but not in a scenario released by companies promoting AV solutions. Finally, Fig. 5(c) shows the realism of BEV provided by VTrackIt compared to the BEV provided by the Level5 dataset. Table 5 shows the diverse range of scenarios presented in the VTrackIt dataset when compared to other state-of-the-art self-driving datasets (both synthetic and real-world). Some salient contributions of the VTrackIt dataset are highlighted in this table.

## IV. CASE STUDY: TRAJECTORY PREDICTIONS

Although the VTrackIt dataset can be used for a wide range of self-driving applications, we focus on trajectory prediction-related tasks in this section. Trajectory prediction is commonly employed in AVs to identify vehicles that present a higher risk to the ego vehicle given its planned trajectory [39]. In their work, Lefèvre *et al.* [39] categorize all trajectory prediction models into three broad categories, namely, (i) Physics-based models [40], [41], (ii) Maneuver-based models [42]–[46] and, (iii) Interaction-aware models [22], [47]–[60]. Today, Graph-based interaction-aware models leveraging semantic maps are considered state-of-the-art [51], [56], [61], [62]. Many state-of-art DL models also use BEV representations to capture complex interactions with other road agents [63]–[68]. Having said that, such models often ignore traffic rules' effect due to limited infrastructure access and surrounding vehicle information. Although many trajectory prediction approaches have been proposed [69]–[74], we propose the first model (InfraGAN) that leverages infrastructure and pooled vehicle information for trajectory predictions. InfraGAN is a two-part network that consists of an interaction-aware trajectory prediction module-

TGAN (Trajectory-GAN) and an LSTM network with a Correction Module (CM).

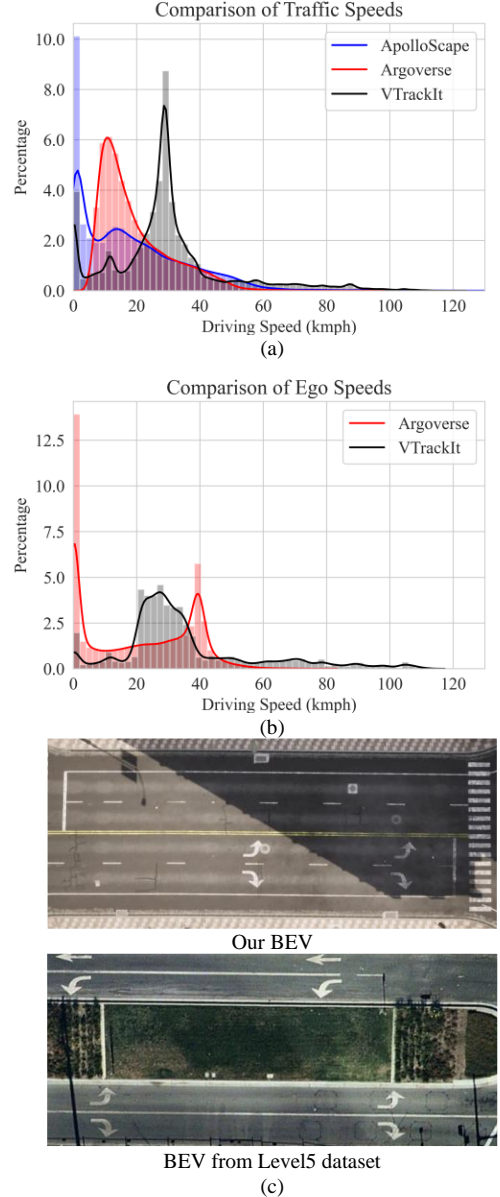

Fig. 5. Comparison of VTrackIt dataset with real-world driving datasets across, (a) traffic speed distributions (for non-stationary vehicles), (b) ego speed distribution (for non-stationary vehicles), and (c) the BEV.

TABLE 5
DIVERSITY OF COMMONLY USED SELF-DRIVING DATASETS

| Dataset | Crashes | Varying Road Surface Conditions | Dynamic Speed Limits | Highway Driving | Traffic Rule Violations | Average Lane Width (m) | Lane Center Annotations | 360° Cameras w/ BEV |
|---|---|---|---|---|---|---|---|---|
| ApolloScape [16] | - | ✓ | - | - | - | 3.84(Miami) | - | - |
| Argoverse [15] | - | ✓ | - | - | - | 3.84(Miami) 3.97(Pittsburgh) | ✓ | - |
| AIODrive [18] | ✓ | - | - | ✓ | ✓ | 3.5 | - | Only 360° views |
| **VTrackIT (Proposed)** | ✓ | ✓ | ✓ | ✓ | ✓ | 3.5 | ✓ (lane deviations are given by 'off_center' in Table 3.) | 360° and BEV |

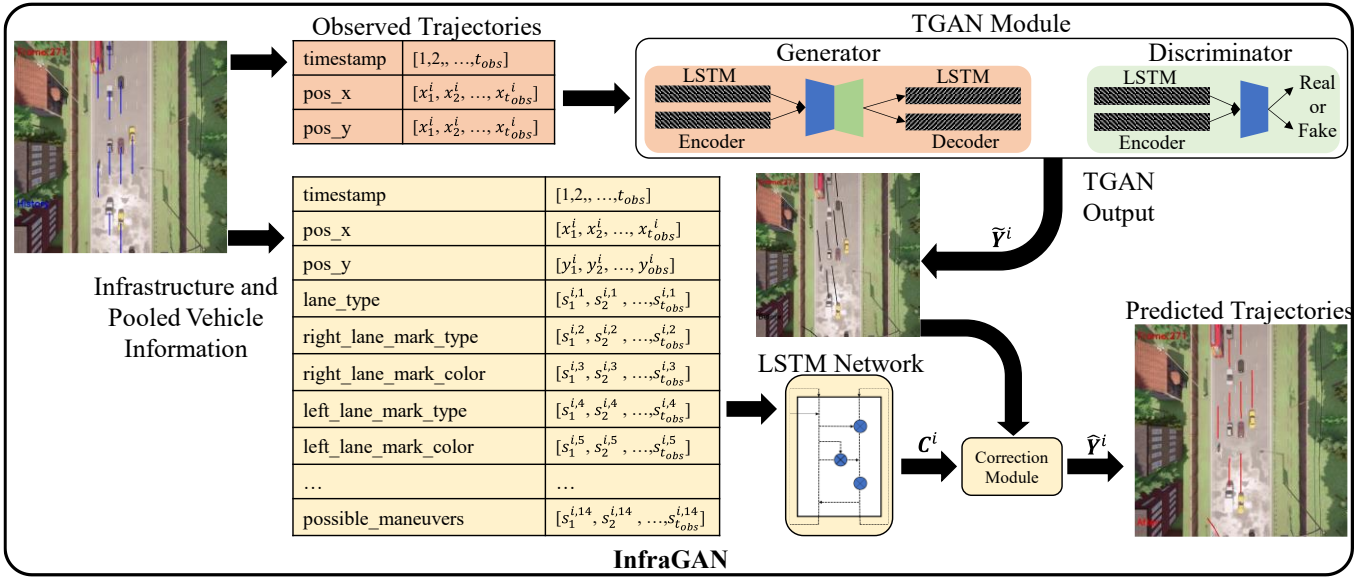

Fig. 6. Overview of the InfraGAN baseline for trajectory prediction using infrastructure and pooled vehicle information in the VTrackIt dataset. The variables $s_t^{i,1}, ..., s_t^{i,14}$ represent the 14 annotated information variables emboldened in Table 3.

## A. InfraGAN: TGAN Module

The TGAN module is a GAN-based DL model that consists of two competing autoencoders (Generator and Discriminator) trained to achieve a zero-sum game where the Discriminator fails to distinguish between generated and training samples. The Generator ($G$) outputs a projected trajectory for a bunch of pooled vehicles $G(z, X)$ for a given random seed $z$ and a given set of historical observed trajectories $X$ for pooled vehicles. The Discriminator ($D$) outputs a probability $D(Y|X)$ that the trajectories $Y|X$ are sampled from the Generator (fake), where $Y$ is the ground truth projected trajectories for the pooled vehicles. The objective of TGAN is to train a robust generator that outputs potential samples of predicted trajectories by solving the min-max problem given in Eq. (1).

$$\min_G \max_D E_{Y|X}[\log D(Y|X)]$$
$$+ E_Z[\log(1 - D(G(z, X)|X))] \qquad (1)$$

Here, $X = [X^1, ..., X^n]$, $n$ is the number of pooled vehicles, including the ego vehicle, $X^i = \left\{(x_t^i, y_t^i)\right\}_{t \le t_{obs}}$ summarizes the observed trajectory of vehicle $i$, $Y = [Y^1, ..., Y^n]$ and $Y^i = \left\{(x_t^i, y_t^i)\right\}_{t_{obs} < t \le t_{obs}+T}$ summarizes the ground truth for future trajectories of the pooled vehicles, $T$ is the number of prediction timesteps, and $z$ is sampled from $\mathcal{N}(0,1)$.

The backbone of the Generator is an LSTM encoder-decoder network. When given an input $X^i$, $G$ outputs a predicted trajectory $\tilde{Y}^i$. The Discriminator then uses this predicted trajectory $\tilde{Y}^i$ or the ground truth trajectory $Y^i$ as input and classifies it as 'Fake' or 'Real.' To standardize the input data, we encode the position of a vehicle $i$ at time $t$ such that

$$e_t^i = \phi(x_t^i, y_t^i; W_{ee}) \qquad (2)$$

Here, $\phi(.)$ represents a single layer, shallow neural network with 'ReLU' non-linearity, and $W_{ee}$ are the weights of the shallow neural network. The LSTM-encoder then uses $e_t^i$ as an input and produces encodings $h_t^{ei}$ for every vehicle $i$ at time $t$ such that

$$h_t^{ei} = \text{LSTM}_e(h_{t-1}^{ei}, e_t^i; W_{encoder}) \qquad (3)$$

Here, $W_{encoder}$ represents the weights of the LSTM encoder ($\text{LSTM}_e$) that are shared between all input vehicles at time $t$ based on suggestions given by Alahi *et al.* [64].

The trajectory predictions can thus be obtained such that

$$(\tilde{x}_t^i, \tilde{y}_t^i) = \gamma(h_t^{di}) \qquad (4)$$

$$h_t^{di} = \text{LSTM}_d(h_{t-1}^{di}, e_t^i; W_{decoder}) \qquad (5)$$

$$e_t^i = \phi(x_{t-1}^i, y_{t-1}^i; W_{de}) \qquad (6)$$

Here, $W_{decoder}$ represents the weights of the LSTM-decoder ($\text{LSTM}_d$), $W_{de}$ represents embedding weights, and $\gamma$ is an MLP. The probability of a trajectory being real/fake is obtained by applying an MLP on the encoder's final hidden state.

The TGAN is trained using both the adversarial loss in Eq. (1) and the mean squared error between the predicted and actual trajectories. While the LSTM-encoder consists of 16 cells, the LSTM-decoder consists of 32 cells. We train the TGAN using a batch size of 64 for 200 epochs using the 'Adam' optimizer with a learning rate of 0.001.

## B. InfraGAN: Correction Module (CM)

The infrastructure and pooled vehicle information (spatial coordinates and 14 additional variables emboldened in Table 3) are given as input to an LSTM network with ten cells and a CM. The LSTM network combined with the CM revises the predicted trajectories from TGAN. For example, the CM tends to eliminate trajectories with lane violations or trajectories that may lead to crashes. The output (predicted) trajectories using InfraGAN are denoted by $\hat{Y}^i$ and mathematically given by the CM as:

$$\hat{Y}^i = \tilde{Y}^i + \tanh(C^i) * \tilde{Y}^i \qquad (7)$$

Here, $\widehat{Y}^i$ is the predicted trajectory for vehicle $i$ using TGAN, $\tanh(C^i)$ is the correction factor for the TGAN predictions using the LSTM network that utilizes the extra information provided by VTrackIt.

The TGAN is first trained for sufficient epochs to produce acceptable trajectories. Then, the entire InfraGAN is trained to correct for TGAN predictions using the 'Adam' optimizer for 200 epochs with an adaptive learning rate. The TGAN is expected to adjust its weights to work in tandem with the LSTM network to achieve reliably corrected trajectories that minimize the loss function of the InfraGAN given in Eq. (8).

$$\mathcal{L}_{\text{InfraGAN}} = \sum_k \left\{ \left\| \widehat{Y}^i - Y^i \right\|_2^2 + \frac{1}{\|C^i\|_1} \right\} \tag{8}$$

In all our experiments, we configure the InfraGAN to generate five likely trajectories ($k = 5$) for every input vehicle $i$ by randomly sampling $z$ from a standard normal distribution. Next, we perform an extensive evaluation using the InfraGAN on the VTrackIt dataset.

### C. Trajectory Prediction Benchmarks

Most trajectory prediction models and AV modules operate at 2-2.5 FPS [51], [55], [65], [75]–[78]. Thus, we down sample the VTrackIt dataset to 2.5 FPS. The scenarios in the VTrackIt 'train' and 'val' sets are then used to train and validate the TGAN and the InfraGAN models. More importantly, to avoid divergence while training the DL model, we exclude scenarios where the ego vehicle had a collision but may still have scenarios where surrounding actors collide. In all our experiments, we observe vehicle trajectory for eight timesteps (or 3.2 secs) and predict trajectory for the following eight timesteps unless mentioned otherwise. We evaluate the baseline methods during testing using the three most used criteria [15], [16], [18]. The Minimum Average Displacement Error (minADE) is the average displacement error for the best-forecasted trajectory for every scenario. The Minimum Final Displacement Error (minFDE) is the final displacement error for the best-forecasted trajectory for every scenario. The Miss Rate is the percentage of sampled forecasted trajectories for every scenario with the final displacement error exceeding 2.0 meters.

While the goal of this work is not to compare baseline models, it is critical to understand how the additional infrastructure and pooled vehicle information affects the trajectory prediction algorithms. In our work, to provide a fair comparison to future models, we provide benchmarks using both- the TGAN and the InfraGAN models. Note that the results reported using the TGAN model do not use any infrastructure and pooled vehicle information apart from the ego and tracked vehicle's $x$ and $y$ coordinates.

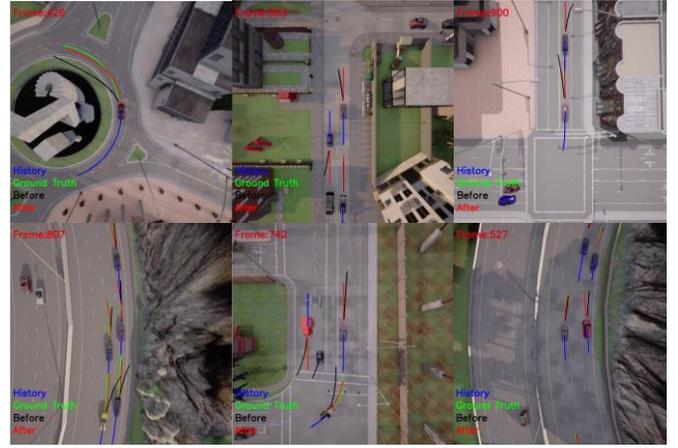

Fig. 7. Performance comparison with and without infrastructure and pooled vehicle information. Observed trajectories are labeled as 'History' and future trajectories are labeled as 'Ground Truth'. Trajectories predicted using TGAN and are labeled as 'Before.' Trajectories predicted using InfraGAN and are labeled as 'After.'

TABLE 6
TRAJECTORY PREDICTION BENCHMARKS USING TOP $k$ PREDICTIONS ON THE VTRACKIT TEST SET

| Metric | $k = 1$ | $k = 3$ | $k = 5$ | Average |
|---|---|---|---|---|
| Without Infrastructure and Pooled Vehicle Information (TGAN) | | | | |
| minADE | 3.19 | 2.11 | 1.87 | 2.39 |
| minFDE | 7.15 | 4.55 | 3.97 | 5.22 |
| Miss Rate | 0.74 | 0.60 | 0.56 | 0.63 |
| With Infrastructure and Pooled Vehicle Information (InfraGAN) | | | | |
| minADE | 1.56 | 1.56 | 1.55 | 1.55 |
| minFDE | 3.58 | 3.57 | 3.56 | 3.56 |
| Miss Rate | 0.43 | 0.42 | 0.42 | 0.42 |

As intuition suggests, the InfraGAN significantly improves over the TGAN by accurately predicting trajectories using the information available in the VTrackIt dataset. This performance gain can be easily visualized in Fig. 7. Table 6 can be used to verify this performance improvement using the InfraGAN over the TGAN when predicting $k$ most likely trajectories using the specified metrics. As is observed in similar GAN-based approaches [65], the TGAN performs significantly worse when put to the test using smaller values for $k$. Further, we also report individualized metrics for every considered map in the test set using Table 7. Table 8 can be used to analyze the performance of both considered models over varying prediction horizons. Note that we always observe the preceding eight timesteps irrespective of the prediction horizon in Table 8. Figures 8 and 9 show the change in the distributions of ADE and FDE for all predicted trajectories using the TGAN and InfraGAN. From these figures, we can clearly see a shorter right tail in the error distributions. Thus, we conclude that the infrastructure and pooled vehicle information provided by VTrackIt supports the development trajectory prediction models that experience a significantly lower number of edge cases.

TABLE 7
COMPARISON OF BENCHMARK MODELS ON THE VTRACKIT TEST SET PER MAP FOR THE TRAJECTORY PREDICTION TASK

| Metric | Town01 | Town03 | Town04 | Town05 | Town06 | Town10 | Overall |
|---|---|---|---|---|---|---|---|
| Without Infrastructure and Pooled Vehicle Information | | | | | | | |
| minADE | 1.50 | 1.93 | 2.74 | **1.13** | 2.45 | 1.62 | 1.87 |
| minFDE | 3.20 | 4.14 | 5.61 | **2.43** | 5.15 | 3.52 | 3.97 |
| Miss Rate | 0.45 | 0.58 | 0.70 | **0.34** | 0.70 | 0.51 | 0.56 |
| With Infrastructure and Pooled Vehicle Information (InfraGAN) | | | | | | | |
| minADE | 1.29 | 1.69 | 2.07 | **0.91** | 2.33 | 1.20 | 1.55 |
| minFDE | 2.93 | 3.89 | 4.75 | **2.15** | 5.34 | 2.75 | 3.56 |
| Miss Rate | 0.32 | 0.44 | 0.51 | **0.25** | 0.59 | 0.35 | 0.42 |

TABLE 8
COMPARISON OF BENCHMARK MODELS ON THE VTRACKIT TEST SET FOR THE TRAJECTORY PREDICTION TASK

| Metric | Pred (2s) 5 Time steps | Pred (2.4s) 6 Time steps | Pred (3.2s) 8 Time steps | Average |
|---|---|---|---|---|
| Without Infrastructure and Pooled Vehicle Information | | | | |
| minADE | 0.63 | 0.95 | 1.87 | 1.17 |
| minFDE | 1.93 | 2.53 | 3.97 | 2.81 |
| Miss rate | 0.36 | 0.44 | 0.56 | 0.54 |
| With Infrastructure and Pooled Vehicle Information (InfraGAN) | | | | |
| minADE | 0.49 | 0.75 | 1.55 | 0.93 |
| minFDE | 1.56 | 2.14 | 3.56 | 2.42 |
| Miss rate | 0.28 | 0.34 | 0.42 | 0.34 |

TABLE 9
EGO TRAJECTORY PREDICTION BENCHMARKS USING TOP $k$ PREDICTIONS ON THE VTRACKIT TEST SET

| Metric | $k = 1$ | $k = 3$ | $k = 5$ | Average |
|---|---|---|---|---|
| Without Infrastructure and Pooled Vehicle Information | | | | |
| minADE | 3.87 | 2.28 | 1.89 | 2.68 |
| minFDE | 8.63 | 4.81 | 3.88 | 5.77 |
| Miss Rate | 0.75 | 0.57 | 0.51 | 0.61 |
| With Infrastructure and Pooled Vehicle Information (InfraGAN) | | | | |
| minADE | 1.80 | 1.79 | 1.78 | 1.79 |
| minFDE | 4.21 | 4.18 | 4.17 | 4.18 |
| Miss Rate | 0.40 | 0.40 | 0.40 | 0.40 |

### D. Ego Trajectory Prediction Benchmarks

Publicly available motion forecasting datasets are also commonly used to build robust motion planning algorithms. As a related task, we use the VTrackIt dataset and utilize recorded data for all tracked vehicles around the ego vehicle while only predicting the trajectory for the ego vehicle. Table 9 can be used to compare the results of the baseline models when predicting $k$ most likely trajectories using mentioned metrics. Using this table, we observe a similar pattern where the InfraGAN outperforms the TGAN. This performance gain can thus be attributed to the additional infrastructure and vehicle information provided in the VTrackIt dataset.

### V. CONCLUSION

In this paper, we introduce and publicly release the VTrackIt dataset. To the best of our knowledge, VTrackIt is the largest, synthetically generated, self-driving dataset annotated with infrastructure and pooled vehicle information variables. This dataset aims to apprise researchers of the benefits of such information in building advanced algorithms for the newer generation of AVs. We also present a compelling case study for trajectory prediction using the VTrackIt dataset. Specifically, we introduce the first trajectory prediction model (InfraGAN) that utilizes infrastructure and pooled vehicle information. Our experiments show that InfraGAN performs better than other methods, resulting in a significantly lower number of edge cases. For future work, we would like to expand this dataset by adding more sensors and challenging scenarios. We would also like to consider additional case studies that integrate and leverage the infrastructure information from the VTrackIt dataset in the near future.

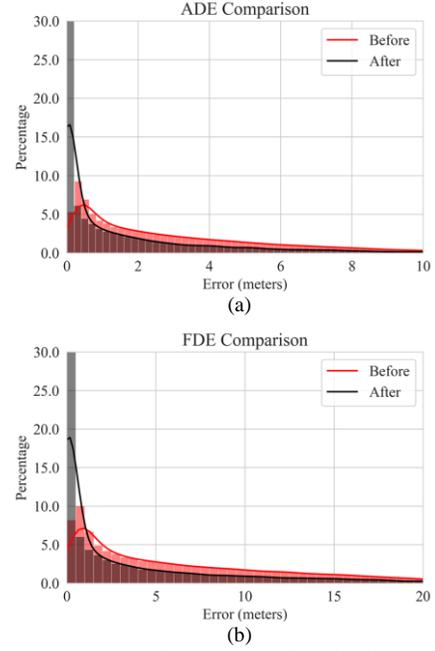

Fig. 8. Comparison of error distributions with and without infrastructure and pooled vehicle information using the VTrackIt test set over – (a) ADE, and (b) FDE. Errors for trajectories predicted using TGAN and are labeled as 'Before'. Errors for trajectories predicted using InfraGAN and are labeled as 'After.'

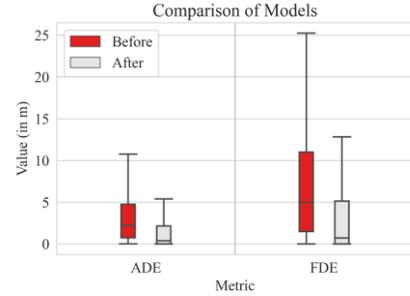

Fig. 9. Comparison of baseline models using the VTrackIt test set. Errors for trajectories predicted using TGAN and are labeled as 'Before'. Errors for trajectories predicted using InfraGAN and are labeled as 'After.'

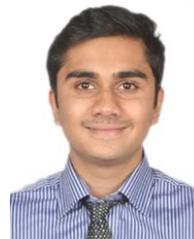

**Mayuresh Savargaonkar** (Student Member, IEEE) received the BS degree in mechanical engineering from the University of Pune, Pune, MH, India, in 2015 and the MS degree in industrial engineering from the University of Michigan-Dearborn, Dearborn, MI, USA, in 2018. He is currently pursuing his Ph.D. in industrial engineering at the University of Michigan-Dearborn, Dearborn, USA. His research interests include autonomous driving, machine vision, and Bayesian methods for machine learning. He is a member of INFORMS, IEEE, and IISE.

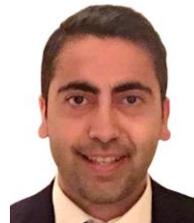

**Abdallah Chehade** (Member, IEEE) received the BS degree in mechanical engineering from the American University of Beirut, Beirut, Lebanon, in 2011 and the MS degree in mechanical engineering, the MS degree in industrial engineering, and the Ph.D. in industrial engineering from the University of Wisconsin-Madison in 2014, 2016, and 2017, respectively. Currently, he is an assistant professor in the Department of Industrial and Manufacturing Systems Engineering at the University of Michigan-Dearborn. His research interests are safe and robust deep learning solutions, data fusion for degradation modeling and prognosis, reliability analytics, and Bayesian statistical modeling. He is a member of INFORMS, IEEE, and IISE.